\documentclass{article}
\usepackage[margin=1.6in]{geometry}
\usepackage{amsmath}
\usepackage{amsfonts}
\usepackage{amssymb}
\usepackage{underscore}
\usepackage{graphicx}
\usepackage{multirow}

\begin{document}

\title{Risk-based Triggering of\\ Bio-inspired Self-Preservation to\\ Protect Robots from Threats}

\author{Sing-Kai Chiu, Dejanira Araiza-Illan, and Kerstin Eder\footnote{Department of Computer Science, University of Bristol, United Kingdom. E-mails: \tt{sc15316.2015@my.bristol.ac.uk, dejanira.araizaillan@bristol.ac.uk, kerstin.eder@bristol.ac.uk}}}
\date{}

\maketitle

\begin{abstract}
Safety in autonomous systems has been mostly studied from a human-centered perspective. 
Besides the loads they may carry, autonomous systems are also valuable property, and self-preservation mechanisms are needed to protect them in the presence of external threats, including malicious robots and antagonistic humans. 
We present a biologically inspired risk-based triggering mechanism to initiate self-preservation strategies.
This mechanism considers environmental and internal system factors to measure the overall risk at any moment in time, to decide whether behaviours such as fleeing or hiding are necessary, or whether the system should continue on its task. 
We integrated our risk-based triggering mechanism into a delivery rover that is being attacked by a drone and evaluated its effectiveness through systematic testing in a simulated environment in Robot Operating System (ROS) and Gazebo, with a variety of different randomly generated conditions. 
We compared the use of the triggering mechanism and different configurations of self-preservation behaviours to not having any of these. 
Our results show that triggering self-preservation increases the distance between the drone and the rover for many of these configurations, and, in some instances, the drone does not catch up with the rover.
Our study demonstrates the benefits of embedding risk awareness and self-preservation into autonomous systems to increase their robustness, and the value of using bio-inspired engineering to find solutions in this area.
\end{abstract}

\section{Introduction}

Autonomous systems such as delivery drones, self-driving cars and robotic assistants are becoming an affordable reality in our daily life. 
Safety aspects so far have been studied from a human-centered perspective, i.e.\ keeping people and people's property safe, exemplified by safety standards for robots that interact and collaborate with people (e.g.\ ISO/TS 15066:2016 Robots and robotic devices -- Collaborative robots).
Nonetheless, as robots and autonomous systems are also valuable property, and so are the loads they carry, they will need to look after their own safety if possible; i.e.\ they will need self-preservation mechanisms in the presence of external threats, such as vandalism and theft~\cite{Bruscic2015,Salvini2010}.

Nature has evolved a range of strategies to survive in a dangerous environment, including morphological, ecological and behavioural adaptations. 
Animals utilize multiple environmental cues to assess whether they are at risk~\cite{Stankowich2005}.
The plasticity to exhibit behaviours in response to a potential threat is crucial for survival. 
Anti-predatory strategies with no detrimental effects on the predator, such as taking refuge, and late resort fleeing mechanisms such a protean flight, provide a source of bio-inspired behaviour for robotic safe threat avoidance, as they ensure safety for both the robot and its antagonist.

Although many strategies such as stealth navigation~\cite{Tews2004} and fleeing behaviours have been designed and implemented for mobile autonomous systems~\cite{Araiza2012,Curiac2015} to avoid dangerous encounters, mechanisms to trigger one or several of these self-preservation strategies to achieve an adequate and timely response to the threats still need to be developed. 
In nature, the instant of evasion initiation depends on many biological and environmental factors~\cite{Cooper2010,Domenici2011}. 
How can we use this knowledge for the design of more competent and fully autonomous systems, able to respond to threats towards robust self-preservation?

In this paper, we propose a novel biologically inspired mechanism that emulates environmental and biological evasion initiation factors, to trigger self-preservation response behaviours based on a risk analysis of the dangerous situation.
We demonstrate the construction and implementation of such a mechanism through a case study consisting of a delivery rover and an attacking drone. 
To evaluate the proposed risk-based triggering mechanism within a cost-effective realistic framework, we implemented a simulator in the Robot Operating System (ROS)~\footnote{http://www.ros.org/} and the 3D physics simulator Gazebo~\footnote{http://gazebosim.org/}. 
In a simulation, the drone pursues the delivery rover either persistently or constrained within a time bound. 
The rover tries to avoid theft or damage by choosing from a variety of predefined response behaviours such as fleeing or seeking refuge, once it has evaluated the risk in the environment in the context of its internal state.

We compared the use of the triggering mechanism and different configurations of response behaviours to not using it at all, i.e.\ a rover that is unaware of the risk and cannot trigger self-preservation responses. 
Our results show that, overall, the triggering mechanism coupled with self-preservation responses has the potential to increase the rover's success on reaching a delivery location, or at least the distance between the threat and the rover. 
This demonstrates the benefits of embedding risk awareness and self-preservation strategies into autonomous systems to increase their robustness, and the usefulness of employing bio-inspired engineering solutions towards achieving true autonomy.

\section{Related Work}

Anti-predator individual mechanisms are divided into different categories: detection avoidance, behavioural vigilance, warning signals, defensive adaptations, and last resort behaviours~\cite{Caro2005}. 
Detection avoidance and defensive adaptations comprise morphological behaviours such as crypsis (matching the background of the environment), weaponry in the body (e.g.\ spines), the release of chemicals~\cite{Barnett2007} to conceal their presence, deceive and mislead predators~\cite{Caro2014}, and also behaviours such as crouching for concealing the body, seeking refuge~\cite{Martin1999}, mobbing and distraction. 
As warning signals, vocal signals warn other animals of predators' presence, whereas displays of coloration advertise potential chemical defence to dissuade predators. 
Morphological adaptations are difficult to implement within the design of robots, although some are emerging, e.g.\ robots that match their background~\cite{Wang2016}. 
Avoidance and defensive behaviours do not negatively impact on the safety of the antagonist.

Last resort behaviours involve increasing the distance between prey and predators. 
Examples are protean behaviour or fleeing away in a zigzag (irregular) manner~\cite{Humphries1970}, along with freezing (immobility) where extreme examples are thanatosis or feigning death, and autotomy or leaving a limb behind. 
Fleeing, freezing and proteanism are well suited to autonomous navigation tasks~\cite{Araiza2012,Curiac2015}.

Animals need to recognize the risk of predation.
Vigilance is a behavioural adaptation where animals alternate between foraging and scanning for potential threats~\cite{Caro2005}. 
Factors and cues such as predator size, approach velocity, perceived sounds, or physical weaponry, influence the choice of response behaviours once a threat has been detected~\cite{Amo2004,Chivers2014,Helfman1989,Smith2001,Stankowich2005,Stankowich2006}. 
As with animals, basic capabilities to assess risk from sensed environmental threats are necessary for robot autonomy.

Risk assessment procedures provide a systematic approach to guide developers in creating autonomous robots that are safe and dependable, from a human-centric perspective--i.e.\ for safe human-robot interactions--, at design time~\cite{MartinG2010,Woodman2012,Dogramadzi2014,Rezazadegan2015}.
Environmental risk analyses can be adopted at runtime, e.g.\ as on-line risk monitors, to control the execution of self-preservation strategies, and even to trigger adaptation and learning towards dealing with threats in the environment as in~\cite{Arcaini2015}. 
These domains, nonetheless, could benefit from considering biologically inspired mechanisms 
for efficient self-preservation responses as well as risk measures and factors. 

This paper proposes such a bio-inspired runtime self-preservation mechanism to trigger different response behaviours according to perceived threats from the environment.
Selecting a response behaviour might mean giving up on other behaviours, such as the delivery of a package, or reaching a final destination, either in the short or the longer term.
The decision to trigger self-preservation behaviours is critical. 
A device is needed to assess whether and when the danger from the environment implies a greater risk and consequently the potential for greater costs and loss, than not reacting to it.

\section{Mechanism to Trigger Self-preservation Behaviours According to Threats}\label{sc:mechanism}

The threats and dangerous situations in the environment that may affect an autonomous system differ widely, depending on the system's application. 
Hazard analysis, as part of a rigorous and systematic risk assessment, involves customers, stakeholders and system designers in the identification and evaluation of the relevant threats and dangers, taking into consideration severity of the harm and the likelihood of it occurring, which results in a risk rating, from low, via moderate and high, to extreme.  
We assume that a set of possible threats has been identified using such a process, and that system designers have equipped the autonomous system with means, including sensing and real-time processing, to detect these in a timely manner.

For example, the analysis in Table~\ref{table:analysis} shows possible generic threats with their risk rating for a delivery rover, according to some hazard analysis, for different types of environments, together with the bio-inspired self-preservation response behaviours to mitigate these, such as fleeing, seeking refuge, thanatosis and autotomy. 
Physical harassment by small animals or children may not pose much of a threat, and adequate responses would include moving away or shutting down for some time (an implementation of thanatosis), in urban environments. 
If the rover is likely to be stolen with its contents, a distraction could be achieved by safely releasing the parcel it carries whilst fleeing (an implementation of autotomy).  
A cross is used to indicate a potentially beneficial response behaviour for the combination of threat and environment.

\begin{table} [!t]
\caption{Analysis of environmental threats and suitable response behaviours in different environments for a delivery rover \label{table:analysis}}
\begin{center}
\begin{tabular}{lllccc}
\hline\noalign{\smallskip}
\multirow{2}{*}{Threats} & \multirow{2}{2cm}{Risk rating} & \multirow{2}{2cm}{Response behaviour} & \multicolumn{3}{c}{Environment} \\
&&& Urban & Open terrain & Indoor\\
\noalign{\smallskip}
\hline
\noalign{\smallskip}
\multirow{4}{2.5cm}{Physical harassment by children or animals} & \multirow{4}{*}{Low} & Fleeing		& 	  & X & \\
														   &						& Seeking refuge		& X & X & X \\
														   &						& Thanatosis			& X & X & X \\
														   &						& Autotomy			& & & \\ \hline \noalign{\smallskip}
\multirow{4}{2.5cm}{Close distance damage} & \multirow{4}{*}{Medium} & Fleeing       & 	  & X & \\
														   &						& Seeking refuge		& X & X & X \\
														   &						& Thanatosis			&  &  &  \\
														   &						& Autotomy			& X & X & \\ \hline \noalign{\smallskip}
\multirow{4}{2.5cm}{Long distance damage} & \multirow{4}{*}{Medium} & Fleeing				& X & X & \\
														   &						& Seeking refuge		& X & X & X \\
														   &						& Thanatosis			&  &  &  \\
														   &						& Autotomy			&  &  & \\ \hline \noalign{\smallskip}
\multirow{4}{2.5cm}{Theft and unauthorized access through physical means} & \multirow{4}{*}{High} & Fleeing&  & X & \\
														   &						& Seeking refuge		& X & X & X \\
														   &						& Thanatosis			&  &  &  \\
														   &						& Autotomy			& X & X & X \\ \hline \noalign{\smallskip}
\multirow{4}{3cm}{Theft and unauthorized access through remote access (hacking)} & \multirow{4}{*}{High} & Fleeing&  & X & \\
														   &						& Seeking refuge		& X & X & X \\
														   &						& Thanatosis			&  &  &  \\
														   &						& Autotomy			& X & X & X \\ 
\hline
\end{tabular}
\end{center}
\end{table}

Qualitative processes to grade the risk of hazards provide metrics to classify their consequences, according to their severity and likelihood of occurrence~\cite{Woodman2012}. 
For example, a risk classification matrix based on the one in the safety standard IEC~61508 `Functional Safety of Electrical/Electronic/Programmable Electronic Safety-related Systems', where four risk classes are possible,
from the most severe (Class I) to the least severe (Class IV), as shown in~\cite{Woodman2012}. 
In our proposed mechanism, we have adapted these qualitative processes to compute a measure to trigger pertinent response behaviours against threats in the environment.

Following an analysis like the one in Table~\ref{table:analysis}, where adequate self-preservation behaviours are chosen as response to particular threats, the next step is the implementation of a mechanism to trigger the start of such responses, once the risk level is assessed and deemed to be at the corresponding level.
We propose the computation of a quantitative measure of risk with respect to the hazards in the environment, and other system-related internal factors that should be accounted for in terms of system safety, the latter emulating internal biological that influence the process of initiating defensive mechanisms in animals.
We consider the existence of $N$ risk factors from environment sensing information collected by an autonomous system, which indicate the type of hazard or threat from the environment towards the system, and hence its risk rating, and $M$ other factors that assess relevant data about the current state of the system (e.g.\ battery life, distance to the destination, proximity to good users). 
Each factor is evaluated through a metric $r_i, i = 1, \ldots , N, N+1,\ldots, N+M, r_i \in \mathbb{R}$, a function over measured or sensed system variables $\bar{x}=[x_1,\ldots,x_j]$ that produces a score, i.e.\ $r_i: \bar{x} \rightarrow \mathbb{R}$. 
An overall risk score $r_{TOTAL}$ can be computed as the (weighted) accumulation of all these $r_i$ factors, e.g.\
\begin{equation}
r_{TOTAL}= \sum_{i=1}^{N+M} w_i\cdot r_i
\end{equation}
to provide a mapping between a level of threat and a response $a \in \mathcal{A}$, i.e.\ $r: \mathbb{R} \rightarrow a$, where $\mathcal{A}$ is the set of all implemented possible response behaviours such as fleeing or freezing (thanatosis).

\section{Case Study}\label{sc:casestudy}

As a case study to evaluate the proposed risk-based self-preservation response triggering mechanism, we continue with the delivery rover example, pursued by an autonomous drone. 
Three particular scenarios from Table~\ref{table:analysis} were employed to create a risk scoring model, for which environmental and internal factors to sense and measure were derived, to compute a risk rating as explained in Section~\ref{sc:mechanism}. 
Additionally, these scenarios were used to choose and implement pre-defined response behaviours to be triggered according to the computed risk, by the response triggering mechanism:
\begin{enumerate}
\item The drone is at a long distance from the rover, where attempts to hack the rover's control towards stealing the delivery consignment can be made. Fleeing has been chosen as the rover's response behaviour by the designer. 
\item The drone is harassing the rover at a closer distance, for which fleeing with proteanism could provide means to confuse the drone.
\item The drone is seeking to damage the rover, approaching until physical contact is made, for which refuge against the drone needs to be sought. 
\end{enumerate}
Note that as the distance between the rover and the drone decreases, the intentions of the drone might become more sinister and the perceived risk of damage to the rover increases accordingly.

After designing the risk scoring model for the triggering mechanism, a simulator was implemented in ROS and Gazebo.
We used available robot models corresponding to real hardware platforms, to provide realism and validity to the experiments, at a computational cost.

\subsection{Instantiation of the Triggering Mechanism for the Case Study}\label{scc:instmechanism}

According to the scenarios, four main environmental and internal factors have been considered for the mechanism to trigger self-preservation responses: the perceived distance between the rover and the drone, the perceived drone sound, the perceived drone speed, and the rover's battery life, i.e.\ $N=3$ and $M=1$. 
Each of these cues is considered to have equal impact in the measured total risk $r_{TOTAL}$.
In practice, different scenarios may require a different weighting of the risk factors, and different number of environmental and internal cues, depending on the environment and what an autonomous system can detect and sense.
The total risk $r_{TOTAL}$ is computed as the accumulation of the relevant individual risk metrics (from the distance $r_d$, sound $r_p$, speed $r_v$ and battery life $r_b$ respectively), each weighted by 0.25,
\begin{equation}
r_{TOTAL} = 0.25r_d + 0.25r_p + 0.25r_v+ 0.25r_b.
\end{equation}

Consider the Euclidean distance between the rover in location $(x,y,z)$ and the drone in location $(x_d,y_d,z_d)$ (all in meters) in the 3D space at time $t$, defined as
\begin{equation}\label{eq:distance}
d(t) = \sqrt{(x - x_d )^2 + (y - y_d )^2 + (z - z_d)^2}.
\end{equation}
We assign a score $s(t)$ that is inverse to the distance $d(t)$, which increases if the drone approaches the rover, and decreases if the rover moves away,
\begin{equation}
s(t)=\frac{100}{d(t)}.
\end{equation}
We then compute five consecutive distance scores, i.e.\ samples $i=1,\ldots,5$ at times $t_1,\ldots,t_5$ (e.g.\ every second).
Consider the gradient of these samples, 
\begin{equation}
\nabla s=\frac{\sum_{i=1}^{5}(s(t_i)-\mu_s)(t_i-\mu_t)}{\sum_{i=1}^{5}(s(t_i)-\mu_s)^2},
\end{equation}
where $\mu_s$ is the average of distance scores over the samples $i=1,\ldots,5$, $\mu_t=3$ (the average of 5 seconds).
If the gradient is positive, the rover is in greater risk of an attack, as the drone has moved closer.
Whereas if the gradient is negative the robot is no longer in as high a risk as it was before. 
Consequently, we propose the computation of the risk given a distance change through the metric 
\begin{equation}
r_d = \left\lbrace \begin{array}{ll}\beta_d \nabla s & \qquad \text{if the gradient is positive} \\ 0 & \qquad \text{if the gradient is negative} \end{array} \right., 
\end{equation}
where $\beta_d$ is a coefficient that normalizes $r_d$ to a value between 0 and 1.

The sound pressure $p$ at time $t$ is calculated from the measured distance $d(t)$ defined in~(\ref{eq:distance}), 
\begin{equation}
p(t) = \frac{60}{d(t)}. 
\end{equation}
The sound pressure increases if the drone gets closer to the rover. 
Note that this measure does not take into account how sound reflects from surfaces, nor the presence of objects in between the origin of sound and the sensor. 
The risk given the sound pressure change is also computed from the gradient of five pressure samples,  
\begin{equation}
r_p=\left\lbrace \begin{array}{ll}\beta_p \frac{\sum_{i=1}^{5}(p(t_i)-\mu_p)(t_i-\mu_t)}{\sum_{i=1}^{5}(p(t_i)-\mu_p)^2} & \qquad \text{if the gradient is positive} \\ 0 & \qquad \text{if the gradient is negative} \end{array} \right.,
\end{equation} 
where $\mu_p$ is the average over the pressure samples, and $\beta_p$ normalizes $r_p$ to a value between 0 and 1.

To calculate an approximation of the relative approach velocity, a sample of the distance $d(t)$ in meters is taken every two seconds (where $d(t_2)$ is the most recent sample, and $d(t_1)$ is the previous sample), and we use the standard definition of the velocity as the difference of the distance over a period of time (in this case 2\,s),
\begin{equation}
v(t)=\frac{d(t_2)-d(t_1)}{2}.
\end{equation}
The risk, given the velocity change, is also computed from the gradient of five approximations,
\begin{equation}
r_v=\left\lbrace \begin{array}{ll}\beta_v \frac{\sum_{i=1}^{5}(v(t_i)-\mu_v)(t_i-\mu_t)}{\sum_{i=1}^{5}(v(t_i)-\mu_v)^2}& \qquad \text{if the gradient is positive} \\ 0 & \qquad \text{if the gradient is negative} \end{array} \right.,
\end{equation} 
where $\mu_v$ is the average over five velocity approximations, and $\beta_v$ normalizes $r_v$ to a value between 0 and 1. 
In general, the velocity of the drone remains constant once the maximum has been reached, with changes only at the initial lift from the ground, and when performing a rotation to face the rover's direction.

Monitoring the battery life is analogous to biological internal factors such as hunger or health status, which influence the kind of triggered anti-predator strategies. 
The remaining battery energy level at time $t$ is computed considering a total capacity of $B_{TOTAL}$, and a linear discharge rate $\phi$, 
\begin{equation}
b(t) = 100-\frac{B_{TOTAL}-t \phi}{6}. 
\end{equation}
The risk given the battery life is calculated according to the energy level,
\begin{equation}
r_b=\beta_b b(t).
\end{equation}
where $\mu_b$ is the average over five computations, and $\beta_b$ normalizes $r_b$ to a value between 0 and 1.

Based on the computed total risk $r_{TOTAL}$, different sets of response behaviours can be programmed, to be triggered when risk thresholds are met. 
For example, the rover decides to pursue its delivery goal if $r_{TOTAL} < \gamma_{flee}$, flee towards the delivery goal if $r_{TOTAL} \geq \gamma_{flee}$, flee with proteanism if $r_{TOTAL} \geq \gamma_{prot}$, or seek refuge if $ r_{TOTAL} \geq \gamma_{ref}$, with $\gamma_{flee} \leq \gamma_{prot} \leq \gamma_{ref}$ as thresholds of risk. 
Alternatively, the rover could perform only one response behaviour, e.g.\ fleeing when $ r_{TOTAL} \geq \gamma_{flee}$.

This risk measuring model reflects the scenarios and the designer's intentions regarding response strategies to avoid financial loss and damage. 
After executing a response behaviour for some time, the total risk $r_{TOTAL}$ is recomputed to determine if a change in behaviour is needed, i.e.\ if the rover should continue with the original task (e.g.\ reaching a delivery goal), try another response behaviour, or continue with the same response, as per the design.

\subsection{Implementation of the Simulator in ROS and Gazebo}

\begin{figure}[t]
\centering
\includegraphics[width=0.7\textwidth]{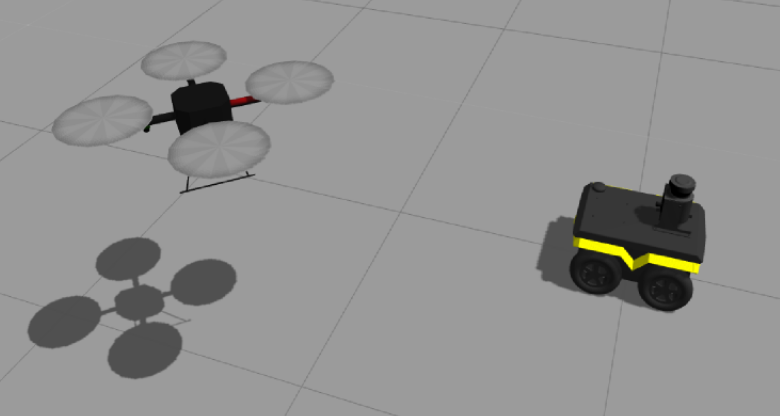}
\caption{Visualization of the 3D simulation of the Jackal rover and the Hector Quadrotor in Gazebo. \label{fig:robots}}
\end{figure}

The ROS framework offers a platform to develop modular software for robots and autonomous systems, consisting of `nodes' (concurrent programs in e.g.\ Python and C++), `topics' (broadcast messages) and `services' (one-to-one communication).
ROS allows distributed computation through a server-client architecture. 
Gazebo is a 3D physics simulator compatible with ROS. 
Many robotic platforms are freely available in simulation for ROS and Gazebo. 
We constructed a simulator that uses the Clearpath Robotics Jackal as the rover\footnote{http://wiki.ros.org/Robots/Jackal}, and the Hector quadrotor\footnote{http://wiki.ros.org/hector_quadrotor} as the drone. 
An example of both robots visualized in a Gazebo simulation is shown in Figure~\ref{fig:robots}. 

\begin{figure}[t]
\centering
\includegraphics[width=0.7\textwidth]{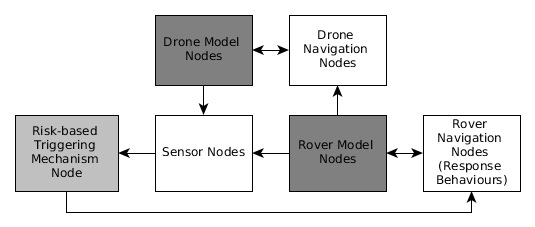}
\caption{Structure of the ROS and Gazebo simulator for the case study, comprising a delivery rover and a drone trying to steal from, or vandalize the rover. A bio-inspired risk-based triggering mechanism selects adequate self-preservation response behaviours in the rover.\label{fig:simulator}}
\end{figure}

The structure of our simulator is shown in Figure~\ref{fig:simulator}. 
The Drone and Rover Model Nodes (in dark gray) comprise the Gazebo 3D models, and the low-level motion control for the actuators (e.g.\ rotation of the wheels).
The implemented bio-inspired risk-based triggering mechanism (as a single node) is shown in light gray, with data inputs from the sensor nodes, and outputs to the Rover Navigation Nodes. 
Other developed nodes (Drone and Rover Navigation and Sensors) are shown in white.
All the developed nodes were implemented in Python. 

The Drone Navigation Nodes control the quadrotor's linear and angular velocity according to readings of the rover's current location in the Gazebo model, aiming to minimize the distance between itself and the rover, $d(t)$. 
The drone indicates if the rover has managed to hide or reach its delivery goal, or if it has been successfully reached, i.e.\ if the distance $d(t)$ is smaller than a minimal threshold, $d_{capture}$. 
The drone rotates over the vertical axis at an angular speed $\omega_d$ to change its orientation towards the rover, and at the linear speed of $v_d$ to pursue the rover. 
A ``persistent'' drone has an infinite amount of battery charge, and will pursue the rover until a drone-rover interaction is finished. 
A ``cautious'' drone considers its finite battery charge, deciding to stop pursuit after some time has elapsed to be able to return to its base safely. 
Notice that the latter mode is more realistic than the former, as it represents a modelling refinement that considers individual costs that the threats in the environment would also need to consider.

The Jackal rover has been programmed to navigate autonomously towards a delivery goal, a location $(x_g,y_g)$, by iteratively performing angle correction at a speed of $\omega$, followed by a linear displacement at a speed $v$. 
If a self-preservation response behaviour is triggered by the risk-based mechanism, the Rover Navigation Nodes execute a combination  of fleeing (moving faster towards the delivery goal), fleeing with proteanism (following sub-goals with randomized orientation angles but avoiding the pursuer, as proposed in~\cite{Araiza2012}), or seeking refuge (navigating towards a refuge in a fixed location), all of these at an increased linear speed of $2v$. 
By decoupling the navigation nodes from the triggering mechanism, the modular structure allows testing different complex self-preservation behaviours.

Sensors Nodes emulate real sensing by reading data from the Gazebo models, such as the location of the drone, and through models of the rover's internal state, such as the state of the battery charge.  
The sensing output is used by the risk-based metrics, embedded in the triggering mechanism node, to trigger adequate response behaviours.

\section{Experiments and Results}

Experiments in simulation were conducted to evaluate a self-preservation triggering mechanism presented in Section~\ref{sc:mechanism}, and instantiated for the case study in Section~\ref{scc:instmechanism}.

\subsection{Setup}

Two different self-preservation configurations were tested in simulation. 
In the configuration A, the rover chooses fleeing, proteanism or seeking refuge, if $r_{TOTAL}$ exceeds the risk thresholds $\gamma_{flee} \leq \gamma_{prot} \leq \gamma_{ref}$, respectively. 
In the configuration B, the rover chooses fleeing if  $ r_{TOTAL} \geq \gamma_{flee}$.
Additionally, the rover does not have a triggering mechanism nor self-preservation behaviours in configuration C. 
Two drone pursuit modes were tested, persistent and cautious, in combination with the three configurations described before.

We generated (pseudorandomly) 150 sets of initial locations for the rover, the drone, the hideaway, and the rover's delivery goal.
The initial locations were restricted so that the distances between the rover and the drone to be sufficiently apart at the start of a simulation. 
Each one of these initial location sets was applied to each configuration A to C, in combination with a persistent or cautious pursuer, for a total of $150 \times 3 \times 2$ simulations. 
This allowed a fair comparison of all the configurations A to C, for the different kinds of pursuers. 
A simulation is run with each set, lasting an allowed maximum of 80 seconds (plus 20 seconds of launching overhead, and 45 of termination). The rover will stop moving if it is reached by the drone, or if it safely reaches the delivery location.

Other setup parameters for the triggering mechanism comprised $\beta_p = \frac{1}{8}$, $\beta_d= \frac{1}{14}$, $\beta_v = \frac{1}{4}$, $\beta_b = \frac{1}{100}$, $B_{TOTAL}=600$, $\phi=1$, $\gamma_{flee}=0.2$, $\gamma_{prot}= 30$, and $\gamma_{ref}=40$. 
For the drone, we used $\omega_d=0.4$\ rad/s, and $v_d=0.5$\ m/s, with a $d_{capture}=0.15$\ m. 
The rover navigates with $v=0.5$\ m/s and $\omega=1.0$ rad/s.

The simulations ran on a PC with Intel 3230M 2.60 GHz CPU, 8 GB of RAM, 64-bit Ubuntu 14.04, ROS Indigo, and Gazebo 2.2.5. 
For each simulation, we collected the sets of initial parameters, type of triggered self-preservation strategy according to $r_{TOTAL}$, and conclusion of the encounter (distances and elapsed simulation time). 
All the logged data and examples of simulations with varied initial conditions and observed behaviours are openly available online~\footnote{https://github.com/riveras/self-preservation}.

\subsection{Results}

We considered the following success criteria during a simulation: reaching the consignment delivery location before capture (strong success); increasing the distance between the drone and the rover when not captured (success); and changing the outcome to reaching the delivery location with configurations A and B, compared to being captured with configuration C, for the same initial condition (relative success). 
We expected that, in general, the first two types of success would be more frequent in the simulations when using the triggering mechanism and the self-preservation behaviours than without using any self-preservation at all.
Using configurations A or B would make the rover reach the delivery goal in instances that it would not without self-preservation.
Furthermore, we expected that using self-preservation would grant more success when the rover was pursued by a cautious drone than with a persistent one, as the rover would have the opportunity to reach the delivery goal once the drone gave up.

\begin{table}[!t]
\centering
\caption{Results with and without the risk-based triggering mechanism, over 150 simulations with different initial conditions.}\label{table:results}
\begin{tabular}{llcccccc}
\hline
&&\multicolumn{5}{c}{Number of simulations} \\ \cline{3-7}
\multicolumn{2}{r}{Configuration} 					& A && B && C\\ 
\multicolumn{2}{r}{Self-preservation behaviour} & All && Fleeing  && None \\\hline
\multicolumn{7}{l}{\textit{Persistent pursuit mode}} \\ \hline
Delivery goal reached (strong success)			&& 116/150	&& 138/150$^\dagger$ 	&& 138/150$^\dagger$\\
Distance increased (success) 					&& 97/118$^\ddagger$	&& 114/138$^\ddagger$ 	&& --\\  
Rover was captured (strong failure)				&& 32/150$^{*}$ 	&& 12/150 	&& 11/150 \\ 
Not captured, goal not reached (inconclusive) 	&& 2/150 	&& 0/150 	&& 1/150\\ 
\scriptsize{Goal reached, out of previously captured with C (rel. success)}	&& 8/11$^\circ$		&& 6/11$^\circ$		&& --\\ 
\scriptsize{Captured, out of previously reaching goal with C (rel. failure)} 				&& 29/138$^\bullet$ 	&& 7/138		&& -- \\\hline 
\multicolumn{7}{l}{\textit{Cautious pursuit mode}} \\ \hline
Delivery goal reached (strong success)				&& 143/150 	&& 145/150$^\dagger$ 	&& 145/150$^\dagger$\\
Distance increased (success) 						&& 87/144$^{\ddagger,\diamond}$ 	&& 60/148$^{\ddagger,\diamond}$ 	&& --\\
Rover was captured (strong failure)					&& 6/150$^{*}$ 	&& 2/150$^\$$ 	&& 5/150 \\
Not captured, goal not reached (inconclusive)		&& 1/150 	&& 3/150 	&& 0/150 \\
\scriptsize{Goal reached, out of previously captured with C (rel. success)}	&& 5/5		&& 5/5$^\#$		&& --\\ 
\scriptsize{Captured, out of previously reaching goal with C	(rel. failure)} 	&& 6/145		&& 2/145$^\#$		&& --\\
\hline
\end{tabular}
\end{table}

Table~\ref{table:results} shows the number of simulations that were successful (according to the success criteria), were inconclusive (i.e.\ by the end of the time limit per simulation, the rover was not captured but did not reach the delivery goal either), failed (i.e.\ the rover was captured), for a drone in two pursuit modes (persistent or cautious), over 150 simulations with different initial locations (for the rover, drone, delivery goal and refuge), and with or without the triggering mechanism and different types of self-preservation behaviours. 
We also recorded which self-preservation strategies were triggered on each simulation, shown in Table~\ref{table:strategies}, to confirm the correct functioning of the triggering mechanism.

The results show that, in general, the combination of the triggering mechanism and only fleeing (configuration B) is more successful than combining the triggering mechanism with the multiple anti-predator behaviours of configuration A, and than not reacting to the threat.
We observed that seeking refuge sometimes lead the rover to move closer to the drone.
Additionally, we observed an oscillation between navigation objectives due to the risk increasing and decreasing: moving towards a refuge or trying to reach the delivery goal, which in some cases caused the rover to be `stuck' in a particular segment of the environment, and the drone was able to get closer. 
These issues are reflected in the strong failure results (see $^*$ in Table~\ref{table:results}).

In terms of the different drone pursuit behaviours, persistent and cautious, the mechanism in configuration B was as strongly successful (i.e.\ it reached the delivery goal) as a rover without any self-preservation (see $^\dagger$ in Table~\ref{table:results}). 
Nonetheless, in terms of increased overall distance between the drone and the rover by the end of a simulation, any of the self-preservation configurations A or B achieved better results for a persistent drone, than for a cautious drone (see $^\ddagger$ in Table~\ref{table:results}), which was contrary to our expectations.
The behaviours in configurations A or B are triggered for longer and at a higher frequency for a persistent drone, which leads to more instances of success than for a cautious drone. 
Furthermore, only fleeing (configuration B) for longer under a persistent drone threat is more efficient at increasing the distance between the rover and the drone, than a combination of self-preservation behaviours (configuration A). 
The opposite happens for a cautious drone, where configuration A outperforms configuration B (see $^\diamond$ in Table~\ref{table:results}). 
This highlights the usefulness of self-preservation behaviours that momentarily change the navigation goals (proteanism or seeking refuge) when the threats in the environment are limited by the management of their own resources.

A rover with configuration A was more successful than one with configuration B at changing the simulation outcomes to reaching the delivery goal for a persistent drone, for the same starting conditions where the rover would be captured with configuration C (see $^\circ$ in Table~\ref{table:results}). 
Nonetheless, new and more capture instances were introduced with configuration A (see $^\bullet$ in Table~\ref{table:results}). 
Only configuration B achieved some relative success, for a cautious drone (see $^\#$ in Table~\ref{table:results}), coupled to the most reduced strong failure results (see $^\$$ in Table~\ref{table:results}).

\begin{table}[!t]
\centering
\caption{Triggering of self-preservation behaviours according to the measured total risk $r_{TOTAL}$ over 150 simulations with different initial locations.}\label{table:strategies}
\begin{tabular}{llrrr}
\hline
&&\multicolumn{3}{c}{Number of simulations} \\ \cline{3-5}
\multicolumn{2}{r}{Configuration} & A && B\\ 
\multicolumn{2}{r}{Self-preservation behaviour}&All  && Only fleeing \\ \hline
\multicolumn{5}{l}{\textit{Persistent pursuit mode}} \\ \hline
Use of simple fleeing && 148/150&  & 148/150 \\
Use of fleeing with proteanism && 59/150&& -- \\
Use of refuge seeking && 25/150&& --\\
No behaviours triggered && 2/150 && 2/150 \\ \hline
\multicolumn{5}{l}{\textit{Cautious pursuit mode}} \\ \hline
Use of simple fleeing && 141/150 && 142/150\\
Use of fleeing with proteanism && 41/150 && -- \\
Use of refuge seeking && 2/150 && --\\
No behaviours triggered && 9/150 && 8/150\\
\hline
\end{tabular}
\end{table}

The results in Table~\ref{table:strategies} show that indeed the triggering of self-preservation behaviours takes place in the majority of the simulations. 
Note that, in the configuration A, different anti-predator behaviours were allowed per simulation. 
Fewer simulations where protean fleeing and seeking refuge were triggered evidence that performing fleeing beforehand helps reducing the risk. 

\subsection{Discussion}

As shown by the results in the previous section, the use of the triggering mechanism in combination with self-preservation behaviours  was successful (i.e.\ increased the distance between the rover and the pursuer in more than half of the simulations with a variety of initial conditions) for a persistent pursuer, and also was strongly successful (i.e.\ allowed the rover to get to the delivery goal in more instances) for a cautious pursuer, compared to not reacting to the threats. 
Nonetheless, particular combinations of self-preservation behaviours were less strongly successful against a persistent pursuer, whereas for a cautious pursuer only fleeing was not that successful. 
Also, relative success results were varied. 
These mixed results, according to our expectations, require further examination of combinations of fleeing and refuge seeking behaviours, to provide a more conclusive evaluation of the triggering mechanism.
Furthermore, anti-predator behaviours coupled with the triggering mechanism should be designed so that they are more effective than `doing nothing'.

An element that influences the functioning of the triggering mechanism is the number and inter-relationships of the risk factors.
Variations of the risk models in Section~\ref{scc:instmechanism}, such as the use of different weights and coefficients, would need to be explored further.
There are evidently trade-offs between avoiding an attack and achieving a successful delivery.
Hence, suitable models and computation of the risk factors need to be explored further, e.g.\ multi-objective optimization. 
Additionally, more sophisticated mechanisms could be used to enhance the risk computation, such as prediction models for the drone.

Threats to the validity of the case study used in this paper and the results include, besides a limited number of combination of self-preservation behaviours and risk factors, the definition of `success' for the evaluation and result reporting.
The selection of some success metrics or criteria over others has an impact on the reported results.
Whereas only considering reaching the delivery goal as `success' is intuitive, it leaves out other aspects of the encounter such as significantly increasing the distance between the drone and the rover, getting outside the line of view of the drone. 
These latter aspects can also be considered as successful encounters from the rover's perspective, and altogether provide a better picture of the effect of the use of the triggering mechanism and the self-preservation behaviours, towards a more holistic evaluation methodology.

\section{Conclusions and Future Work}

We presented a biologically inspired risk-based triggering mechanism to initiate self-preservation strategies.
This mechanism considers environmental and internal system factors to measure the overall risk at any moment in time, to decide whether behaviours such as fleeing or hiding are necessary, or whether the system should continue with its task. 
This emulates animal anti-predator behaviour initiation. 
The mechanism's design is based on risk assessment methodologies for robotics design, complementing traditional human-centered safety analyses towards systems with more autonomy and self-preservation.

A case study was developed to evaluate such a triggering mechanism coupled with different self-preservation strategies, compared against not reacting to threats. 
In the case study, a delivery rover is attacked by a drone in a simulated environment in ROS and Gazebo, with a variety of different randomly generated conditions such as initial locations, and delivery goals.

Our study demonstrates the need for embedding risk awareness and self-preservation towards successful autonomous systems, and the usefulness of bio-inspired engineering solutions. 
In general, the triggering mechanism coupled with self-preservation strategies increases the distance between the threat of the drone and the rover. 
Nonetheless, some of the self-preservation behaviours lower the frequency of reaching the delivery goal.

As future work, an extensive study of combinations of adequate and optimized self-preservation behaviours is necessary to determine what actions lead to achieving a delivery objective while increasing the distance between the treat and the rover. 
Additionally, new risk metrics that consider more complex factors such as probable future actions (i.e.\ prediction) for the threats could be incorporated into the mechanism to obtain a more robust risk measure.

\bigskip\noindent{\bf Acknowledgement:} 
The work by D. Araiza-Illan and K. Eder was funded by the EPSRC project ``Robust Integrated Verification of Autonomous Systems'' (ref. EP/J01205X/1).

\bibliographystyle{plain}

\bibliography{references}

\end{document}